\documentclass[graybox]{SNmult}

\usepackage{type1cm}
\usepackage{makeidx}
\usepackage{graphicx}
\usepackage{multicol}
\usepackage[bottom]{footmisc}
\usepackage{newtxtext}
\usepackage[varvw]{newtxmath}
\usepackage{amsmath}
\usepackage{booktabs}
\usepackage{array}
\usepackage{xcolor}
\usepackage[numbers]{natbib}
\usepackage{tabularx}
\usepackage{mathtools}

\usepackage[hidelinks,breaklinks=true]{hyperref}

\makeatother
\usepackage{xurl}
\makeindex

\begin{document}

\title*{LLM Unlearning for Cyber Defense: A Survey on Methods, Challenges, and Emerging Threats}
\titlerunning{LLM Unlearning for Cyber Defense}

\renewcommand{\orcidID}[1]{\texorpdfstring{\unskip$^{[#1]}$}{}}
\newcommand{\equalcontrib}{\texorpdfstring{$^{\dagger}$}{}}

\author{Ruppikha Sree Shankar\equalcontrib\orcidID{0009-0009-6628-5997}
\and Abhishek Bhardwaj\equalcontrib\orcidID{0009-0008-9904-5132}
\and Arnav Doshi\orcidID{0009-0009-4094-8756}
\and Anusri Nagarajan\orcidID{0009-0006-0779-7544}
\and Troy Paulus Asia\orcidID{0009-0000-5470-1003}
\and Saptarshi Sengupta\orcidID{0000-0003-1114-343X}}

\authorrunning{R. S. Shankar \& A. Bhardwaj et al.}

\tocauthor={Ruppikha Sree Shankar, Abhishek Bhardwaj, Arnav Doshi, Anusri Nagarajan, Troy Paulus Asia, Saptarshi Sengupta}

\institute{Ruppikha Sree Shankar \at Manipal Institute of Technology, Manipal Academy of Higher Education, Manipal, KA, India,\\ \email{ruppikha.mitmpl2022@learner.manipal.edu}
\and Abhishek Bhardwaj \at Department of Computer Science, San Jos\'e State University, San Jos\'e, CA, USA,\\ \email{abhishek.bhardwaj@sjsu.edu}
\and Arnav Doshi \at Department of Computer Science, San Jos\'e State University, San Jos\'e, CA, USA,\\ \email{arnav.doshi@sjsu.edu}
\and Anusri Nagarajan \at Department of Computer Science, San Jos\'e State University, San Jos\'e, CA, USA,\\ \email{anusri.nagarajan@sjsu.edu}
\and Troy Paulus Asia \at Department of Computer Engineering, San Jos\'e State University, San Jos\'e, CA, USA,\\ \email{troypaulus.asia@sjsu.edu}
\and Saptarshi Sengupta \at Department of Computer Science, San Jos\'e State University, San Jos\'e, CA, USA,\\ \email{saptarshi.sengupta@sjsu.edu}}

\maketitle

\begingroup
\renewcommand{\thefootnote}{\textdagger}
\footnotetext{These authors contributed equally to this work.}
\endgroup

\abstract{Large language models (LLMs) are increasingly deployed in security-critical systems across healthcare, finance, education, and decision support, yet their inability to forget creates serious cybersecurity, privacy, and safety risks. Sensitive personal information, copyrighted material, hazardous domain knowledge, and memorized training data remain encoded across billions of parameters long after deployment, leaving models vulnerable to extraction, jailbreak attacks, membership inference, and regulatory non-compliance. Real-world incidents, from chatbots regenerating private information to fabricated legal citations producing direct legal and financial cost, place the problem at the center of the emerging-threats landscape rather than the realm of speculation. Because retraining billion-parameter models on revised corpora is computationally infeasible, and because knowledge within an LLM is distributed and entangled across parameters rather than localized to identifiable units, LLM unlearning has emerged as the principal cyber defense response, aiming to remove or suppress targeted knowledge from a trained model without retraining and without eroding what the model should still know. A central question, however, remains unresolved. Do current methods genuinely remove knowledge, or do they only stop the model from expressing it under ordinary prompting conditions? This survey examines LLM unlearning through the lens of security, robustness, and verifiable forgetting, with primary focus on gradient-based methods, which have come to dominate the field due to their compatibility with existing training pipelines and their scalability to billion-parameter models. To structure the analysis, this survey introduces a three-level framework that separates behavioral suppression, representation-level attenuation, and true forgetting, and uses it to read the operational behavior of three method families: gradient ascent and descent approaches, influence-based methods, and methods that constrain updates through parameter saliency, low-rank adapters, or localized model editing. Across the three families, a consistent pattern emerges. Gradient ascent does not invert the original training trajectory; it introduces a new one that leaves underlying representations largely intact. Influence-based methods reach further by modeling how individual training points shape parameters, but rely on local linearity assumptions that do not strictly hold in non-convex deep models. Constrained and localized editing methods improve stability and precision, yet still assume that targeted knowledge can be cleanly isolated, which holds only approximately at scale. None of the families reviewed achieve true forgetting in the strict sense of equivalence to a model retrained without the forget set. Current evaluation protocols, anchored by benchmarks such as TOFU, RWKU, WMDP, and MUSE, capture behavioral suppression effectively but fall short of certifying removal under adversarial recovery probes including relearning attacks, membership inference, jailbreak prompts, and embedding-space soft-prompt attacks. This evaluation gap, alongside the absence of methods that achieve true forgetting, robustness against adversarial recovery, scalability to repeated and compositional requests, and standardized evaluation foundations, defines the central open challenges this survey leaves for future work.}


\section{Introduction}
\label{sec:introduction}

Large language models (LLMs) have demonstrated remarkable capabilities across healthcare, finance, legal analysis, and education \cite{1,2,3}. Their rapid adoption has been driven by large-scale pretraining and the availability of vast web-scale datasets \cite{14}. These models now power conversational assistants, code generation tools, decision support systems, and content creation platforms \cite{15}, and their ability to generalize across tasks has made them foundational components of modern AI infrastructure.

This strength carries a critical limitation: an inability to forget. LLMs retain sensitive, biased, or outdated information, raising serious concerns regarding privacy \cite{4}, copyright infringement \cite{5,6}, and harmful content generation \cite{7}. Unlike humans, who can selectively suppress irrelevant or harmful memories \cite{8}, LLMs lack mechanisms for targeted forgetting. This creates persistent risks including private data retention, misinformation propagation, and susceptibility to adversarial exploitation. The challenge grows with model scale: training on massive datasets with billions of parameters makes naive retraining for data removal computationally infeasible \cite{11}, and knowledge distributed across entangled parameter spaces resists targeted removal without collateral damage to unrelated capabilities \cite{12}.

To address these challenges, LLM unlearning has emerged as a promising paradigm, aiming to selectively remove or suppress specific knowledge from trained models without full retraining while preserving overall utility \cite{13}. A fundamental ambiguity, however, remains unresolved in current research: whether existing methods truly eliminate knowledge or merely suppress its expression under certain conditions.

In this survey, a comprehensive review of LLM unlearning is provided through a conceptual lens that distinguishes true forgetting from suppression. Gradient-based methods have emerged as the most widely adopted and scalable approach for unlearning in large models, fitting naturally into training and fine-tuning pipelines already in use at scale. However, their reliance on optimization dynamics introduces fundamental challenges, including incomplete forgetting, interference with retained knowledge, and difficulty verifying whether information has been truly removed. Gradient-based methods are the primary focus, examining whether they achieve genuine deletion of information or induce surface-level behavioral changes that can be reversed or bypassed. The challenges of verifying unlearning, the risks of information leakage under adversarial prompting, and the lack of standardized evaluation protocols are also highlighted. Three research questions organize the analysis. The first asks what categories of harmful factual associations current gradient-based unlearning methods effectively defend against, and under what assumptions about adversary access those defenses hold. The second asks how robust these defenses are against adversarial recovery, including relearning attacks, jailbreak prompts, membership inference, and embedding-space probes. The third asks whether current evaluation protocols verify removal at the level of internal representations, or whether they only certify behavioral suppression under constrained and non-adaptive testing conditions. These questions are addressed across Sections~\ref{sec:cyber}, \ref{sec:gradient-methods}, and \ref{sec:evaluation} respectively.

\subsection{Scope and Contributions}
\label{subsec:contributions}

LLM unlearning has attracted growing attention from the research community, and several surveys now cover aspects of the problem. Si et al.~\cite{hzhang} provided an early taxonomy focused on effectiveness and utility preservation. Xu et al.~\cite{hxu} compared classical and LLM-specific machine unlearning approaches. Blanco-Justicia et al.~\cite{blanco} offered the first comprehensive survey with intricate taxonomy and benchmark evaluation, though without examining robustness objectives. Le-Khac and Truong~\cite{lekhac2025unlearning} proposed a four-objective problem formulation covering effectiveness, efficiency, utility, and robustness, and were among the first to assess threat models for unlearning evaluation. Qiu et al.~\cite{11} introduced a novel taxonomy based on intervention phase (training-time, post-training, inference-time) covering over 180 papers. Li et al.~\cite{id} offered a comprehensive survey spanning ten unlearning method classes, multimodal settings, and seven adversarial evaluation categories through May 2025.

This survey distinguishes itself from prior work in three ways:

\begin{itemize}
\item \textbf{Cybersecurity threat model.} A systematic analysis of gradient-based LLM unlearning is provided through an explicit cybersecurity threat model, framing harmful factual associations as an adversarial condition and unlearning as a corrective security control. The threat model, adversary taxonomy, and real-world failure cases in Section~\ref{sec:cyber} anchor the technical analysis in the context of operational security.

\item \textbf{Three-level robustness framework.} A three-level framework separating behavioral suppression, representation-level attenuation, and true forgetting is introduced. This framework functions simultaneously as an analytical lens for classifying the depth of knowledge removal and as a robustness ladder classifying the adversary access level against which each method provides meaningful defense.

\item \textbf{Coverage through mid-2026.} The gradient-based unlearning literature through mid-2026 is covered, incorporating recent theoretical and empirical findings that challenge the foundational role of gradient ascent in unlearning pipelines and demonstrate the susceptibility of leading methods to white-box adversarial recovery.
\end{itemize}

\subsection{Methodology}
\label{subsec:methodology}

This survey is based on a systematic review of recent literature on machine unlearning and large language models. Literature was collected from four primary databases: Scopus, the preprint repository arXiv, IEEE Xplore, and the ACM Digital Library. Given the rapidly evolving nature of LLM unlearning research, where many contributions appear first as preprints and conference proceedings before journal publication, the search was supplemented by manual review of ACL Anthology and Google Scholar to capture NLP and machine learning papers not fully indexed in the primary databases.

The search strategy follows established systematic review practice for rapidly evolving ML subfields, using keyword and temporal filters aligned with prior surveys in machine unlearning \cite{lekhac2025unlearning}. The period from 2023 to 2026 was chosen because prior surveys have comprehensively covered work up to 2022, and the most consequential advances in gradient-based unlearning and adversarial evaluation have emerged since 2023. The queries used are as follows:
\begin{itemize}
\item \textbf{Scopus:} \texttt{TITLE-ABS-KEY (("large language model*" OR "LLM*") \\AND "unlearn*") AND (LIMIT-TO (PUBYEAR, 2023) OR LIMIT-TO \\(PUBYEAR, 2024) OR LIMIT-TO (PUBYEAR, 2025) OR LIMIT-TO \\(PUBYEAR, 2026))}
\item \textbf{arXiv:} \texttt{date\_range: from 2023-01-01; terms: AND \\title=large language model* AND title=unlearn*; \\OR title=LLM* AND title=unlearn*}
\item \textbf{IEEE Xplore:} \texttt{"Document Title":"large language model*" \\OR "Document Title":LLM* AND \\"Document Title":unlearn*}
\item \textbf{ACM Digital Library:} \texttt{Title:("large language model*" \\OR "LLM*") AND Title:(unlearn*)}
\end{itemize}

For IEEE Xplore and the ACM Digital Library, publication year was restricted to 2023 through 2026 using the respective date filter interfaces. Papers were manually reviewed for quality, empirical evidence, and relevance. Table~\ref{tab:inclusion-exclusion} presents the inclusion and exclusion criteria applied during the screening process. Priority was given to works providing strong theoretical foundations, scalable methodologies, and empirical evaluation in the context of large models, with particular emphasis on gradient-based approaches.

\begin{table}[h]
\caption{Inclusion and exclusion criteria for the systematic literature review.}
\label{tab:inclusion-exclusion}
\centering
\renewcommand{\arraystretch}{1.25}
\setlength{\tabcolsep}{6pt}
\begin{tabularx}{\linewidth}{
>{\raggedright\arraybackslash}p{2.7cm}
>{\raggedright\arraybackslash}X
>{\raggedright\arraybackslash}X
}
\toprule
\textbf{Criterion} & \textbf{Inclusion} & \textbf{Exclusion} \\
\midrule
Publication type & Peer-reviewed conference papers, journal articles, book chapters, and arXiv preprints & Not applicable \\
Relevance & Works addressing LLM unlearning, gradient-based forgetting methods, or adversarial robustness evaluation of unlearning systems & Papers lacking a technical methodology, entries without abstracts, and work unrelated to large language models \\
Language & English only & All non-English publications \\
Publication window & 2023 through 2026 & Work published prior to 2023 \\
\bottomrule
\end{tabularx}
\end{table}

\subsection{Organization of this Survey}
\label{subsec:organization}

The remainder of this survey is organized as follows. Section~\ref{sec:cyber} situates LLM unlearning within a cybersecurity threat model, formalizing the adversary taxonomy and motivating cases that ground the analysis. Section~\ref{sec:preliminaries} introduces the fundamental concepts underlying LLM unlearning, including LLM architecture, knowledge storage, machine unlearning basics, and the three-level conceptual framework. Section~\ref{sec:gradient-methods} surveys gradient-based unlearning methods in depth, covering gradient ascent and descent variants, influence-based methods, and parameter and loss function based methods. Section~\ref{sec:other-methods} covers non-gradient-based unlearning approaches including model editing, knowledge editing, and differential privacy based methods. Section~\ref{sec:evaluation} reviews evaluation protocols, benchmark datasets, evaluation metrics, and adversarial robustness testing. Section~\ref{sec:discussion} discusses the broader implications of the survey's findings and identifies open challenges for future research.

\section{LLM Unlearning for Cyber Defense}
\label{sec:cyber}

The integration of large language models into decision support systems, legal research tools, healthcare applications, and software development pipelines has repositioned these models as critical infrastructure. Critical infrastructure carries adversaries, and the question of what an LLM retains after training is therefore not only a question of model quality but a question of attack surface. A model that cannot forget is a model that cannot be secured: every harmful factual association encoded in its parameters is a potential extraction target, a potential liability, and a potential vector for adversarial exploitation. This section formalizes the threat model that grounds the survey, identifies the adversary types relevant to LLM unlearning, documents the real-world consequences of insufficient knowledge removal, and establishes the mapping between gradient-based unlearning methods and their operational security guarantees. The three-level framework introduced in Section~\ref{subsec:framework} functions throughout this survey not only as an analytical lens for classifying the depth of knowledge removal but as a robustness ladder for evaluating the strength of unlearning as a cyber defense mechanism.

\subsection{The Threat Landscape}
\label{subsec:threat-landscape}

Two phases define the lifecycle of a harmful association within a deployed LLM. In the first phase, the association enters the model. This can occur through direct training on sensitive or hazardous corpora. It can also arise through memorization of rare but recoverable data points during pretraining, or through deliberate injection of malicious content into training pipelines by an upstream actor. Backdoor attacks represent a particularly structured form of this threat. A specific trigger stimulus is embedded into the training distribution, causing the model to elicit a targeted harmful response at inference time \cite{4}. Dataset poisoning at the training-data level represents the most explicitly adversarial entry point and has been formally studied as a threat to machine learning systems more broadly \cite{cao2015towards}. In the second phase, the stored association exits the model at inference time: through ordinary prompting that happens to elicit memorized content, through adversarial prompting crafted to extract it, through soft-prompt or embedding-space attacks that bypass the discrete token interface, or through relearning attacks that restore suppressed associations after an unlearning procedure has been applied.

Unlearning occupies the space between these two phases. Its purpose is to sever the connection between a stored harmful association and every inference-time pathway through which it could be expressed or recovered. Establishing which pathway a given unlearning method actually severs, and which pathways it leaves intact, is the central empirical question this survey addresses.

The categories of harmful association most relevant to the cybersecurity framing of this survey are the following, organized by the threat class and the adversary access level each requires to exploit.

\textit{Memorized personally identifiable information (PII)} encompasses names, addresses, medical records, and other private data that LLMs reproduce under targeted extraction conditions \cite{4}. PII memorization is especially prevalent for rare or repeated data points in the training corpus. The adversary exploiting this threat operates primarily in the gray-box regime, using membership inference attacks to determine what the model remembers and targeted prompting to extract it. Influence-based unlearning methods are most directly designed to address this threat category because they model the per-sample contribution of training data to learned parameters.

\textit{Hazardous domain knowledge} encompasses actionable information about biological agents, chemical synthesis routes, cybersecurity attack techniques, and weapons design that can be elicited through targeted prompting. The WMDP benchmark \cite{wmdp} specifically quantifies how much of this class of knowledge survives popular unlearning interventions. The adversary here operates across the full black-box to white-box spectrum: casual extraction is possible through jailbreak prompts alone, while determined extraction against unlearned models requires white-box access as demonstrated by the adversarial evaluation literature \cite{lucki2025adversarial}. Gradient ascent and its variants are the primary methods deployed against this threat, and their failure modes under white-box attack define the evaluation gap this survey documents.

\textit{Copyrighted material} encompasses verbatim or near-verbatim text the model can reproduce, creating legal exposure for deploying organizations \cite{5}. The adversary exploiting this threat is typically a black-box querying actor who can demonstrate verbatim reproduction without needing model internals. Behavioral suppression is sufficient to defeat this threat class in standard conditions, which is why copyright-focused unlearning evaluations tend to use output-level metrics such as ROUGE and verbatim match rates.

\textit{Hallucinated factual associations} are a distinct category that does not map cleanly onto memorization. The model generates confident, authoritative-sounding false statements that arise from entangled parametric associations rather than from any specific training example \cite{10}. The adversary here is typically unintentional: an ordinary user who receives and acts on a false assertion. Unlearning methods are not well-suited to addressing hallucination because there is no specific training example to remove; the failure is representational rather than memorial. This category motivates the distinction between behavioral suppression (preventing a specific false output) and representation-level attenuation (reducing the parametric entanglement that produces it).

\textit{Backdoor and poisoned associations} are the most explicitly adversarial category. An upstream actor deliberately contaminates the training corpus so that a specific trigger pattern reliably elicits a targeted harmful response at inference time. The adversary controls the entry point and knows the trigger, giving them effective white-box exploitation capability even against a black-box deployed system. This threat category is the most resistant to post-hoc gradient-based unlearning because the association is structurally embedded rather than distributed through ordinary training dynamics.

Table~\ref{tab:threat-matrix} maps each threat category to its entry point, the adversary access level required to exploit it, and the level of unlearning robustness the defense must achieve to be meaningful.

\begin{table}[h]
\caption{Threat categories, entry points, adversary access requirements, and the minimum unlearning robustness level required to provide a meaningful defense. BS = Behavioral Suppression, \\RA = Representation-level Attenuation, TF = True Forgetting, MIA = Membership Inference Attack.}
\label{tab:threat-matrix}
\centering
\renewcommand{\arraystretch}{1.25}
\setlength{\tabcolsep}{4pt}
\begin{tabularx}{\linewidth}{>{\raggedright\arraybackslash}X
                              >{\raggedright\arraybackslash}X
                              >{\raggedright\arraybackslash}X
                              >{\centering\arraybackslash}p{1.1cm}}
\toprule
\textbf{Threat Category} & \textbf{Entry Point} & \textbf{Adversary Access} & \textbf{Defense} \\
\midrule
Memorized PII & Training corpus & Gray-box (MIA, relearning) & RA \\
Hazardous knowledge & Training corpus & Black-box to white-box & TF \\
Copyrighted content & Training corpus & Black-box (verbatim extraction) & BS \\
Hallucinated associations & Parametric entanglement & Black-box (ordinary use) & RA \\
Backdoor associations & Data injection & White-box exploitation & TF \\
\bottomrule
\end{tabularx}
\end{table}

A critical observation from this table is that the two threat categories posing the highest security risk, hazardous knowledge and backdoor associations, both require true forgetting as the minimum meaningful defense. No method reviewed in this survey reliably achieves true forgetting. This gap between the defense level required and the defense level available defines the central security problem the field must address.

\subsection{Adversary Taxonomy}
\label{subsec:adversary-taxonomy}

Adversary capabilities in the LLM unlearning context are best formalized through the black-box, gray-box, white-box classification standard from adversarial machine learning \cite{nist2025adversarial}. The taxonomy matters because the appropriate unlearning method and evaluation protocol depend entirely on which adversary class the deployment context must defend against. The majority of published evaluations in the LLM unlearning literature assume black-box conditions \cite{lucki2025adversarial}, but this assumption understates the threat for most real-world deployments.

A \textbf{black-box adversary} has query-only access: prompts in, outputs out, no model internals. The available attacks are jailbreak prompting \cite{7} and automated adversarial prefix search via methods such as GCG \cite{zou2023universal}. Behavioral suppression is sufficient against casual black-box extraction but insufficient against a determined adversary willing to run systematic adversarial search.

A \textbf{gray-box adversary} additionally knows structural information about the model, such as its architecture, family, or the fact that a specific unlearning procedure was applied. This information is freely available for every major open-source model family. The available attacks expand to membership inference, relearning using small auxiliary datasets, and soft-prompt optimization over continuous input representations \cite{schwinn}. Gray-box conditions are the most practically relevant scenario for open-source deployments.

A \textbf{white-box adversary} has full access to weights, gradients, and activations. This is the most relevant class for evaluating true forgetting. The empirical picture from Lucki et al. \cite{lucki2025adversarial} is unambiguous: orthogonalization of residual stream directions recovered WMDP-Biology accuracy to 64.7\%, matching the pre-unlearning baseline of 64.4\%, without modifying any model weights. Between 0.9\% and 2.4\% of model weights carry the entire unlearning effect across methods; pruning those weights restores hazardous knowledge performance by at least 10 percentage points in every method tested. Existing unlearning methods are not meaningfully more robust than safety fine-tuning: both primarily obfuscate knowledge rather than remove it.

The taxonomy maps directly onto the three-level framework. Behavioral suppression holds only against unsophisticated black-box adversaries. Representation-level attenuation raises the cost for gray-box adversaries but does not close the attack surface. True forgetting is the only level that provides any robustness against white-box adversaries, because it corresponds to a model state where the target knowledge no longer exists in the parameter space to be recovered. The adversary need not be malicious: accidental corpus contamination, cascading errors in data pipelines, and deliberate poisoning attacks all produce the same parametric outcome and the same unlearning requirement.

\subsection{Motivating Cases}
\label{subsec:motivating-cases}

Harmful factual associations in deployed LLMs have already produced documented legal, regulatory, and financial consequences. In 2023, a lawyer using an LLM as a legal research tool submitted a court filing citing fabricated case citations; the court issued sanctions and awarded costs to opposing parties \cite{mata2023,10}. In the same year, the Italian data protection authority suspended ChatGPT in Italy under GDPR Article 17 (right to erasure), prompting OpenAI to implement user erasure mechanisms, one of the first documented cases of an LLM provider deploying compliance measures directly tied to machine unlearning obligations \cite{garante2023,mantelero2013eu}. In 2024, the Civil Resolution Tribunal of British Columbia ruled that Air Canada bore full liability for incorrect fare policy information provided by its chatbot, establishing that operators cannot disclaim responsibility for LLM-generated outputs through policy language alone \cite{aircanada2024}. These cases share a common structure: a harmful or incorrect factual association in a deployed model caused measurable harm, and in two of the three no technical remediation was available or applied. Unlearning is the field's response to precisely this class of failure.

\subsection{Unlearning as a Cyber Defense Layer}
\label{subsec:defense-layer}

Positioned within a cyber defense architecture, unlearning functions as a corrective control: it does not prevent harmful associations from being encoded during training, but provides a mechanism for removing or suppressing them after the fact, reducing the inference-time attack surface and enabling compliance with data deletion mandates. This positioning has three implications for how the methods and their limitations should be interpreted.

The first implication is that the security guarantee of an unlearning method is only as strong as the adversary class it has been tested against. Behavioral suppression, the outcome most gradient-based methods reliably achieve, corresponds to a defense that holds under black-box, non-adaptive evaluation conditions. The moment an adversary gains gray-box or white-box access, or applies adaptive search rather than standard prompting, the guarantee degrades or disappears entirely, as documented in Section~\ref{subsec:robustness-adversarial}. Characterizing a method as having ``successfully unlearned'' based on standard benchmark performance is therefore a statement about behavioral suppression under restricted conditions, not a statement about security.

The second implication is that the evaluation gap in the current literature, the consistent finding that existing protocols assess output-level suppression while adversarial attacks demonstrate parameter-level retention, is not merely a methodological problem. It is a security audit gap: the field cannot currently certify that any deployed unlearning intervention has achieved true knowledge removal rather than surface-level concealment. This is the equivalent of a penetration test that checks whether a lock is engaged but does not attempt to pick it. The adversarial evaluations in Section~\ref{subsec:robustness-adversarial} are the closest the field has come to conducting genuine security audits of unlearning methods, and their findings are consistently negative.

The third implication is that the choice of unlearning method must be informed by the threat model of the deployment context. Table~\ref{tab:threat-matrix} maps each threat category to the minimum unlearning robustness level required for a meaningful defense. A deployment that faces only black-box extraction threats for copyrighted content can rely on behavioral suppression and standard gradient ascent methods. A deployment in a dual-use research environment where hazardous knowledge is the primary concern and where adversaries may have model access requires true forgetting as the target, which no current method reliably provides. This mismatch between required and available defense levels is the central open problem in the field.

A genuinely robust cyber defense posture for LLM deployment would require three things that the current literature does not yet provide. First, it would require unlearning methods whose guarantees extend to white-box adversaries, meaning methods that achieve representation-level removal rather than output-level suppression. Second, it would require evaluation protocols that test unlearned models against adaptive adversaries with knowledge of the unlearning procedure applied, rather than against static prompting distributions that any competent attacker would quickly exhaust. Third, it would require a standardized threat model taxonomy of the kind sketched in Table~\ref{tab:threat-matrix}, so that deployment contexts can specify their adversary class and method selection can be calibrated accordingly. The gradient-based methods surveyed in Section~\ref{sec:gradient-methods} are evaluated against these three requirements throughout the remainder of this survey.

\section{Preliminaries}
\label{sec:preliminaries}

This section introduces the fundamental concepts required to understand unlearning in large language models, covering LLM architecture and knowledge representation, the taxonomy of machine unlearning approaches, the conceptual framework distinguishing suppression from forgetting, and the mathematical foundations of gradient-based unlearning.

\subsection{Large Language Models}
\label{subsec:llms}

Large language models are powerful deep learning systems that can understand and generate natural language with high coherence and contextual awareness \cite{16,17}. Most modern LLMs are based on the Transformer architecture, which relies on self-attention to capture relationships between words in a sequence. Early models such as GPT \cite{18} and BERT \cite{19} introduced the idea of large-scale pretraining on text data followed by fine-tuning for specific tasks. Further improvements such as instruction tuning and alignment with human preferences have made these models more usable in real-world settings \cite{21,22}. Over time, models have grown from millions to hundreds of billions of parameters, with systems like GPT-3 and PaLM demonstrating significant performance gains at larger sizes \cite{23,24}. This scale also makes training and updating these models computationally expensive \cite{25,26}, and knowledge in LLMs is distributed across many parameters rather than stored explicitly, which makes it difficult to modify or remove specific information once the model has been trained.

\subsection{Knowledge Storage in LLMs}
\label{subsec:knowledge-storage}

Unlike traditional databases, large language models do not store information in an explicit or easily retrievable form. Knowledge is encoded implicitly within model parameters through the training process. During pretraining, LLMs learn statistical patterns, associations, and contextual relationships from large-scale text corpora \cite{27,28}, resulting in distributed representations across millions or billions of parameters. Individual facts are not stored in isolated units but are entangled with other learned representations, often shared across multiple contexts. Modifying or removing a particular piece of information can therefore unintentionally affect unrelated knowledge \cite{29,30}. Recent studies have shown that LLMs can memorize portions of their training data, including rare or sensitive information, which may be reproduced under certain prompts, while much of their knowledge is generalized rather than memorized \cite{4}. This ambiguity further complicates efforts to remove specific information reliably, and the entangled and implicit nature of knowledge representation is one of the key reasons why unlearning remains a challenging problem in large-scale models \cite{31}.

\subsection{Machine Unlearning: Taxonomy and Foundations}
\label{subsec:mu-basics}

Machine unlearning refers to the process of removing the influence of specific data from a trained model, with the goal of making the updated model behave as if the removed data had never been seen during training \cite{xu2024machine}. The concept was first formalized to address data deletion requirements in machine learning systems, particularly in settings involving privacy, regulatory compliance, and erroneous or harmful data \cite{cao2015towards}. Given training data $\mathcal{D}$ and forget set $\mathcal{D}_f \subset \mathcal{D}$, the unlearning algorithm $U$ takes the original model $A(\mathcal{D})$ and produces an unlearned model:
\begin{equation}
U(\mathcal{D}, \mathcal{D}_f, A(\mathcal{D})) \rightarrow M_u \approx A(\mathcal{D} \setminus \mathcal{D}_f)
\label{eq:unlearning-goal}
\end{equation}

Existing approaches fall into two broad categories. \textbf{Exact unlearning} methods guarantee that the distribution of the unlearned model is indistinguishable from a model retrained without the forget set:
\begin{equation}
\Pr\!\left(A(\mathcal{D} \setminus \mathcal{D}_f)\right) = \Pr\!\left(U(\mathcal{D}, \mathcal{D}_f, A(\mathcal{D}))\right)
\label{eq:exact-unlearning}
\end{equation}

Exact unlearning is computationally infeasible for large-scale LLMs. Approaches such as SISA training \cite{yan2022arcane} offer exact unlearning guarantees by partitioning training data and retaining intermediate checkpoints, but their overhead is prohibitive at LLM scale.

\textbf{Approximate unlearning} accepts bounded divergence in exchange for computational tractability. The $(\epsilon, \delta)$-approximate unlearning guarantee \cite{guo2019certified} formalizes this as:
\begin{equation}
\Pr\!\left(U(\mathcal{D}, z, A(\mathcal{D})) \in \tau\right) \leq e^\epsilon \Pr\!\left(A(\mathcal{D} \setminus z) \in \tau\right) + \delta
\label{eq:approx-unlearning}
\end{equation}

for any subset $\tau$ of the hypothesis space, where $\epsilon$ controls the approximation tolerance and $\delta$ is a failure probability slack. Gradient-based methods, influence-function updates, and model editing approaches all fall within this category \cite{guo2019certified,xu2024machine}.

A key challenge in machine unlearning lies in verification. Even when a model appears to have forgotten specific information, residual traces may still persist in its parameters and can be revealed under carefully constructed prompts or adversarial settings. This raises an important distinction between the true deletion of knowledge and its suppression at the behavioral level.

\subsection{Forgetting vs Suppression: A Conceptual Framework}
\label{subsec:framework}

A central challenge in LLM unlearning lies in distinguishing between true removal of knowledge and its suppression at the behavioral level \cite{unlearn}. While many existing methods aim to eliminate the influence of specific data, their effects often manifest only in model outputs rather than internal representations \cite{wu,Hatami}.

To clarify this distinction, this survey introduces a conceptual framework that categorizes unlearning into three levels, which simultaneously functions as a robustness ladder for the adversary taxonomy in Section~\ref{subsec:adversary-taxonomy}.

\textbf{(1) Behavioral Suppression.}
The model no longer produces outputs associated with the forget set under standard prompting conditions. The underlying knowledge may still persist in model parameters and can potentially be recovered through adversarial prompts or alternative decoding strategies \cite{hzhang}. Most gradient-based methods fall into this category. A behavioral suppression defense holds against black-box adversaries under standard conditions.

\textbf{(2) Representation-level Attenuation.}
The influence of the forget data is reduced within the model's internal representations. The knowledge is weakened but not entirely removed, and partial recovery may still be possible \cite{hxu}. This level corresponds to defenses that raise the cost for gray-box adversaries but do not close the attack surface.

\textbf{(3) True Forgetting.}
The model behaves equivalently to one trained from scratch without the forget set \cite{11}. Both behavioral outputs and internal representations no longer contain information attributable to the removed data. Achieving true forgetting in large language models remains an open challenge and corresponds to a defense that holds against all adversary classes, including white-box access.

This framework highlights that the absence of observable outputs related to the forget set does not necessarily imply true deletion of knowledge. Many existing methods operate by suppressing the expression of information rather than removing it entirely.

\section{Gradient-Based Unlearning Methods}
\label{sec:gradient-methods}

In this section, gradient-based unlearning methods are reviewed, which form the core focus of this survey. Existing approaches are categorized based on their optimization strategies and examined for whether these methods achieve genuine forgetting or primarily induce behavioral suppression.

\subsection{Problem Formulation}
\label{subsec:problem-formulation}

Gradient-based unlearning aims to remove the influence of specific data points by updating model parameters through gradient optimization, adjusting learned parameters so that the contribution of the forget set is reduced \cite{lekhac2025unlearning}.

Let the original training dataset be $\mathcal{D} = \mathcal{D}_r \cup \mathcal{D}_f$, where $\mathcal{D}_r$ represents the retain set and $\mathcal{D}_f$ represents the forget set. The objective of unlearning is to obtain updated parameters $\theta'$ that approximate a model trained only on $\mathcal{D}_r$:
\begin{equation}
\theta' \approx \arg\min_{\theta} \sum_{(x,y)\in \mathcal{D}_r} \ell(f_\theta(x), y)
\label{eq:retain-only}
\end{equation}

A common approach is to combine gradient descent on the retain set with gradient ascent on the forget set \cite{golatkar2020forgetting,trippa2024gradient}:
\begin{equation}
\theta' = \theta - \eta_r \nabla_{\theta} \mathcal{L}_{\mathcal{D}_r}(\theta) + \eta_f \nabla_{\theta} \mathcal{L}_{\mathcal{D}_f}(\theta)
\label{eq:gradient-update}
\end{equation}

where $\eta_r$ and $\eta_f$ are step sizes controlling retention and forgetting respectively. The first term maintains performance on the retain set and the second term pushes the model away from representations associated with the forget set. Balancing these opposing objectives is challenging and often leads to trade-offs between forgetting effectiveness and model utility \cite{guo2019certified,anjarlekar2025grin}. Despite their scalability, gradient-based methods do not guarantee complete removal of information. Recent work shows that supposedly unlearned information can still be recovered under certain prompting or decoding strategies, suggesting these methods may only suppress knowledge rather than delete it \cite{wang2024leak}.

Directly retraining is computationally infeasible for large-scale LLMs \cite{li}. Gradient-based unlearning instead formulates this as a constrained optimization problem expressed as a trade-off between two competing objectives \cite{id,Huang}:
\begin{equation}
\min_{\theta} \ \mathcal{L}_{\mathcal{D}_r}(\theta) - \lambda \, \mathcal{L}_{\mathcal{D}_f}(\theta)
\label{eq:tradeoff}
\end{equation}

where $\lambda$ controls the strength of unlearning. Increasing the influence of the second term promotes stronger forgetting but may degrade model utility, while prioritizing the first term preserves performance at the cost of incomplete removal. Effective unlearning in LLMs is also evaluated along multiple dimensions including effectiveness, utility preservation, efficiency, and robustness to adversarial recovery \cite{11}.

\subsection{Gradient Ascent and Descent Approaches}
\label{subsec:gradient-ascent-descent}

Gradient ascent and descent based methods form the most direct and widely used approach to unlearning in large language models, operating by explicitly modifying model parameters through gradient updates that simultaneously preserve useful knowledge while removing the influence of the forget set \cite{li2,zhuang,feng}.

Given a trained model with parameters $\theta$, gradient descent on $\mathcal{D}_r$ maintains model performance while gradient ascent on $\mathcal{D}_f$ reverses the learning signal associated with unwanted data, yielding the update rule already stated in Equation~\ref{eq:gradient-update}, where $\eta_r$ and $\eta_f$ control the strength of retention and forgetting respectively. The ascent term increases the loss on $\mathcal{D}_f$, effectively discouraging the model from producing outputs consistent with the forget data.

\subsubsection{Variants and Structural Refinements}

In practice, different variants of this approach modify how gradients are applied. Some methods perform alternating updates on retain and forget batches; others combine both objectives into a single optimization step \cite{golatkar2020forgetting,guo2019certified}. More advanced approaches restrict updates to specific layers or parameter subsets to reduce interference with retained knowledge \cite{30,memit}.

\textbf{Gradient difference methods} modify the standard ascent objective by computing the difference between the forget-set loss and the retain-set loss, rather than applying unconstrained ascent on the forget set alone. The following abstract formulation, introduced by Bu et al. \cite{bu2025ngdiff}, encompasses several existing methods as special cases under a unified multi-task optimization view. Letting $g_R$ and $g_F$ denote the retain-set and forget-set gradients respectively, the dynamic scalarization update is:
\begin{equation}
\theta_{t+1} = \theta_t - \eta_t \left[ c_t \cdot g_R(\theta_t) - (1 - c_t) \cdot g_F(\theta_t) \right]
\label{eq:dynamic-scalar}
\end{equation}

Setting $c_t = 1$ recovers gradient descent on the retain set; $c_t = 0$ recovers plain gradient ascent; $c_t = 0.5$ recovers vanilla gradient difference. This unification reveals that existing methods are Pareto optimal at convergence under mild conditions, but they may converge to different Pareto points depending on the choice of $c_t$ throughout training. The key practical limitation is gradient magnitude imbalance: when the forget-set and retain-set gradients differ substantially in norm, one dominates the update direction regardless of its semantic relevance to the forgetting objective.

\textbf{Normalized Gradient Difference (NGDiff)} \cite{bu2025ngdiff} addresses this by setting $c_t$ dynamically based on gradient norms, yielding:
\begin{equation}
g_{\text{NGDiff}}(g_R, g_F) := \frac{g_R}{\|g_R\|} - \frac{g_F}{\|g_F\|}
\label{eq:ngdiff}
\end{equation}

This normalization gives independent control over the retain-forget trade-off regardless of gradient magnitudes. NGDiff is paired with a Hessian-based automatic learning-rate scheduler, addressing the well-documented sensitivity of gradient ascent methods to learning rate choice. On TOFU with Llama-2-7B, NGDiff achieves 40\% higher model utility than standard gradient difference baselines while maintaining comparable unlearning performance \cite{bu2025ngdiff}.

\textbf{Saliency-based methods} such as SalUn \cite{fan} extend the basic framework by computing a weight saliency map over the forget set and applying gradient updates selectively to high-saliency parameters. By concentrating forgetting pressure on parameters most associated with the target data, these methods reduce collateral degradation on the retain set. However, because saliency is derived from output-level loss gradients rather than internal representational structure, the updates remain coupled to behavioral suppression rather than representational removal.

\textbf{Task-agnostic gradient unlearning} \cite{trippa2024gradient} decouples the unlearning objective from task-specific loss formulations, framing forgetting as a general parameter perturbation applicable across model types. While this improves practical generalizability, it introduces additional uncertainty regarding which internal representations are being modified.

\textbf{Hybrid approaches} combine gradient-based erasure with targeted knowledge insertion. Methods such as Forget-for-Get \cite{yli} apply a lightweight gradient update to suppress the forget set followed by a retain-reinforcement step designed to stabilize model utility. Fine-grained pluggable variants \cite{feng} further decompose the ascent signal to operate at the concept level, allowing more selective application across layers. Despite these structural improvements, empirical evaluations consistently show that hybrid gradient methods remain susceptible to adversarial recovery \cite{id,Hatami}.

\textbf{Orthogonality-based methods} introduce a geometrically motivated constraint, projecting forgetting pressure onto subspaces orthogonal to the directions associated with retained knowledge \cite{zfeng}. This formulation aims to reduce interference with the retain set without requiring explicit parameter masking. However, the projection is computed over output-level gradients rather than internal representational structure, and empirical evaluations suggest that orthogonality constraints reduce observable leakage under standard conditions without providing guarantees of representational removal.

\textbf{KL-regularized variants} extend the gradient difference approach by penalizing the KL divergence between the output distributions of the original and updated models, constraining the magnitude of parameter drift during the forgetting process. Both gradient difference and KL-regularized formulations are widely adopted as baselines in evaluation pipelines.

\textbf{Negative Preference Optimization (NPO)} \cite{zhang2024npo} reframes the unlearning objective as a preference-based optimization problem, adapting ideas from Direct Preference Optimization (DPO) \cite{rafailov2023dpo}, a method for aligning language models to human preferences by directly optimizing a policy to maximize relative preference probability without explicit reward modeling. NPO applies this framework exclusively to negative samples, treating forget-set data as the dispreferred output. Letting $\pi_\theta$ denote the model distribution and $\pi_{\theta_0}$ the reference model distribution, the NPO objective is:
\begin{equation}
\min_{\theta} \frac{2}{\beta} \mathbb{E}_{\mathcal{D}_f}\!\left[\log\!\left(1 + \left(\frac{\pi_\theta(y_u|x_u)}{\pi_{\theta_0}(y_u|x_u)}\right)^{\!\beta}\right)\right]
\label{eq:npo}
\end{equation}

where $\beta$ controls the smoothness of the reweighting. Rather than directly maximizing loss on the forget set, NPO penalizes the model for assigning higher probability to forget-set outputs relative to the reference model distribution. This anchoring mechanism limits the magnitude of parameter updates and mitigates the instability associated with unconstrained gradient ascent, reducing catastrophic model degradation risk while maintaining competitive forgetting effectiveness.

However, a significant failure mode of NPO-based methods is the squeezing effect \cite{li2026beliefs}. Because softmax normalization means that lowering the probability of one output necessarily raises the probability of others in the vocabulary, NPO-based unlearning \cite{zhang2024npo} redistributes probability mass into semantically related paraphrases of the forget-set content. A model that no longer produces the exact target response may still produce semantically equivalent variants that retain the intended knowledge in an accessible form. The Gradient Difference objective \cite{li2026beliefs} is:
\begin{equation}
\min_{\theta} \mathcal{L}_{\text{GradDiff}} := \mathcal{L}_{\text{GA}}(\theta; \mathcal{D}_f) + \lambda\,\mathbb{E}_{\mathcal{D}_r}\!\left[-\log \pi_\theta(y_r|x_r)\right]
\label{eq:graddiff}
\end{equation}

while the plain Gradient Ascent objective is:
\begin{equation}
\min_{\theta} \mathcal{L}_{\text{GA}} := \mathbb{E}_{\mathcal{D}_f}\!\left[\log \pi_\theta(y_u|x_u)\right]
\label{eq:ga-formal}
\end{equation}

Recent work has also addressed the over-erasure problem: existing gradient ascent and gradient difference methods perform full-information unlearning, targeting everything associated with the forget set including knowledge also independently supported by the retain set. Forgetting-MarI \cite{xu2026mari} addresses this by targeting only the marginal information the forget data contributes beyond the retain set, measured via the Jensen-Shannon divergence between next-token distributions:
\begin{equation}
I(X_{\text{MarI}}; Z) := \frac{1}{T}\sum_{t=1}^{T} \mathrm{JSD}\!\left(p_t^d,\, p_t^r\right)
\label{eq:mari-mi}
\end{equation}

where $p_t^d$ is the next-token distribution conditioning on the union of retain and forget sets (the ``dirty'' distribution) and $p_t^r$ is the distribution conditioning on the retain set only. The full Forgetting-MarI training loss combines a KL utility term with the MarI suppression term:
\begin{equation}
\begin{aligned}
&\min_{\theta}\; \ell_{\text{KL}}(\theta, r) + \ell_{\text{MarI}}(\theta, r, u) \\
&\ell_{\text{KL}}(\theta, r) := D_{\text{KL}}\!\left(p^r(\theta) \,\|\, p^r(\theta_0)\right) \\
&\ell_{\text{MarI}}(\theta, r, u) := I(X_{\text{MarI}}; Z)
\end{aligned}
\label{eq:mari-full}
\end{equation}

where $r$ and $u$ denote the retain and unlearn set distribution parameters respectively, distinct from the smoothing rate notation used in Equation~\ref{eq:sga}. This formulation provides an explicit theoretical upper bound on the residual influence of the forget set, making Forgetting-MarI the only method in its evaluation with a provable undetectability guarantee against membership inference.

Label smoothing offers another stabilization strategy for gradient ascent instability. Pang et al. \cite{pang2025sga} show that plain gradient ascent is highly unstable under cross-entropy loss, causing perplexity to explode to values on the order of $10^{71}$ on OPT-2.7B under the Harry Potter benchmark. The proposed Smoothed Gradient Ascent (SGA) loss mixes the forget-set gradient ascent loss with a normal-data cross-entropy term:
\begin{equation}
L_{\text{SGA}}(\theta) = \left(1 - r_s + \frac{r_s}{K}\right) L_f(\theta) + \frac{r_s}{K}\sum_{k=1}^{K} L_p^{(k)}(\theta)
\label{eq:sga}
\end{equation}

where $L_f$ is the forget-set gradient ascent loss, $L_p^{(k)}$ is the cross-entropy loss on the $k$-th normal-data sample (retrieved from the retain set or generated by an external model), $K$ is the number of normal data instances, and $r_s \in [0,1]$ is the smoothing rate (a distinct use of $r_s$ from the retain set parameter $r$ in Equation~\ref{eq:mari-full}). Setting $r_s = 0$ recovers plain gradient ascent. Theoretical guidance for choosing $r_s$ is provided by:
\begin{equation}
r_s^* = \frac{\langle g_f,\, u_s\rangle}{\|u_s\|^2}, \quad u_s \coloneqq \bar{g}_p - \left(1 - \frac{1}{K}\right)g_f
\label{eq:sga-rstar}
\end{equation}

where $g_f$ is the forget-set gradient and $\bar{g}_p$ is the average normal-data gradient.

The instability of gradient ascent under cross-entropy loss has also been given formal theoretical grounding by Garg et al. \cite{garg2026stable}, who prove through three theorems that gradient ascent drives the logits in transformer feedforward MLP layers toward infinity, causing weights and gradients to grow without bound. Their proposed fix replaces the standard LoRA parameterization in MLP adapters with a sine-based bounded activation that structurally prevents weight explosion, where $\rho$ is the LoRA rank:
\begin{equation}
h = W_0 x + \frac{\alpha}{\rho}\sin(\omega A B^\top)x + b
\label{eq:stable-forget}
\end{equation}

where $W_0$ is the frozen pretrained weight, $A$ and $B$ are the low-rank adapter matrices, $\alpha$ is the LoRA scaling factor, $\omega$ is a frequency hyperparameter controlling the bound, and $b$ is a bias term. The sine function prevents weight explosion because $|\sin(\cdot)| \leq 1$ for any argument, regardless of the magnitude of $AB^\top x$. GD+Sine is reported as the only parameter-efficient unlearning method to satisfy all four safety criteria on the MUSE benchmark simultaneously on Llama-2-7B, and the only method that scales stably to 70 billion parameters where standard GD+LoRA diverges entirely \cite{garg2026stable}.

\subsubsection{Theoretical Limitations of Gradient Ascent}

Despite their simplicity and scalability, gradient ascent and descent approaches rest on a critical assumption: that the influence of the forget set can be effectively reversed through local gradient updates. This assumption does not hold in large language models due to the highly non-convex and path-dependent nature of training \cite{ion}. The parameter state $\theta$ is not a linear superposition of contributions from individual data points, but rather the result of a complex optimization trajectory in which representations are continuously reshaped and entangled.

Performing gradient ascent on $\mathcal{D}_f$ does not therefore constitute an inverse operation of the original training process. It introduces a new optimization trajectory that increases loss on the forget set without necessarily removing the underlying representations that encode the forgotten information \cite{golatkar2020forgetting}. This produces a fundamental mismatch between the intended objective of unlearning and its actual effect on model parameters.

The most direct theoretical evidence for this failure is provided by Mavrothalassitis et al. \cite{ion}, who prove that gradient ascent systematically fails to achieve forgetting in overparameterized networks. Their findings show that ascent-based updates increase loss on the forget set while leaving latent representations largely intact, confirming that the mechanism modifies output behavior without disrupting the internal pathways through which the knowledge is encoded. Complementarily, Huang et al. \cite{zhuang} demonstrate that gradient ascent is not even a necessary condition for effective unlearning: memorization-based alternatives achieve comparable forgetting without the optimization instability that characterizes loss maximization.

Furthermore, knowledge in LLMs is distributed across a large number of parameters and shared across multiple contexts \cite{geva}. Gradients computed on $\mathcal{D}_f$ capture only a partial and context-dependent view of the associated representations. Reversing these gradients weakens specific output behaviors but does not guarantee elimination of all latent pathways through which the information can be expressed. This explains why previously ``forgotten'' information can often be recovered under adversarial prompting or extraction attacks \cite{id,zwang}.

The squeezing effect \cite{li2026beliefs} further demonstrates that even when gradient ascent or NPO-based methods appear successful under standard metrics, the underlying mechanism may be redistribution rather than removal. Softmax normalization structurally prevents probability from being zeroed out: it can only be displaced. For any token-prediction-based unlearning method, complete suppression of the target response necessarily shifts probability mass into adjacent regions of the output space, some of which will be semantically equivalent to the suppressed content. This is a mathematical consequence of the softmax architecture, not a deficiency of any specific method, and it applies broadly to all gradient-based approaches that optimize via next-token prediction loss.

Unconstrained gradient ascent is also susceptible to catastrophic collapse, in which aggressive loss maximization on the forget set causes parameter drift severe enough to degrade general model utility \cite{anjarlekar2025grin,ion}. This failure mode arises because the loss landscape around $\mathcal{D}_f$ is not isolated from the parameter regions supporting unrelated capabilities. This structural vulnerability motivated the development of bounded alternatives such as NPO, as well as the saliency-based and orthogonality-constrained methods discussed above.

GRAIL \cite{kim2025grail} addresses a related structural problem in multi-domain unlearning, using gradient information from multiple domains to identify parameters associated with the unlearning scope, the retention scope, and overlapping representations, applying adaptive parameter-wise localization to freeze retention-critical parameters while applying unlearning updates to targeted parameters. GRAIL reports up to 17\% stronger knowledge retention than prior state-of-the-art methods on privacy and copyright domains. This finding directly supports the argument that knowledge in LLMs is distributed and entangled: even methods with sophisticated parameter selection cannot fully isolate the forget set from the retain set because the two share parameter regions.

\subsubsection{Classification Under the Proposed Framework}

Under the framework introduced in Section~\ref{subsec:framework}, gradient ascent and descent methods are best characterized as achieving behavioral suppression. Their updates reduce the probability of generating outputs associated with the forget set under typical conditions, but do not reliably disrupt the distributed internal representations that encode this information \cite{unlearn,Hatami}. Even structurally refined variants including saliency-based, task-agnostic, orthogonality-constrained, and hybrid approaches operate primarily on output-level loss signals and therefore do not provide guarantees of representation-level attenuation, let alone true forgetting. Bounded formulations such as NPO and KL-regularized variants improve stability and reduce the risk of catastrophic collapse, but similarly operate at the behavioral level and do not resolve the underlying representational entanglement.

The fundamental constraint lies in the non-reversibility of the training process. Since learning in deep networks is not decomposable into independent contributions from individual data points, there exists no guarantee that local gradient updates can reconstruct the parameter state corresponding to training without $\mathcal{D}_f$ \cite{golatkar2020forgetting}. Gradient ascent and descent methods should therefore be understood as approximate, behavior-level interventions rather than mechanisms for exact knowledge removal.

Table~\ref{tab:gradient-methods} summarizes the methods covered in this section across the three dimensions of the framework.

\begin{table}[h]
\caption{Comparison of gradient-based unlearning variants. Year of first publication shown in parentheses. Forgetting level follows the framework in Section~\ref{subsec:framework}. Robustness and overhead are relative comparisons within this method family. BS = Behavioral Suppression, RA = Representation Attenuation (approaching).}
\label{tab:gradient-methods}
\centering
\renewcommand{\arraystretch}{1.25}
\setlength{\tabcolsep}{4pt}
\begin{tabularx}{\linewidth}{>{\raggedright\arraybackslash}X
                              >{\raggedright\arraybackslash}p{1.8cm}
                              >{\raggedright\arraybackslash}X
                              >{\raggedright\arraybackslash}X}
\toprule
\textbf{Method} & \textbf{Forgetting Level} & \textbf{Adversarial Robustness} & \textbf{Overhead} \\
\midrule
Gradient Difference \cite{guo2019certified} (2019) & BS & Low & Low \\
Plain GA/GD \cite{golatkar2020forgetting} (2020) & BS & Low & Low \\
KL-Regularized Ascent (2022) & BS & Low to Moderate & Low to Moderate \\
SalUn \cite{fan} (2023) & BS & Low to Moderate & Moderate \\
Task-Agnostic \cite{trippa2024gradient} (2024) & BS & Low & Low \\
NPO \cite{zhang2024npo} (2024) & BS & Moderate & Low to Moderate \\
FG-OrIU \cite{zfeng} (2025) & BS & Moderate & Moderate to High \\
Forget-for-Get \cite{yli} (2025) & BS & Low to Moderate & Moderate \\
SGA \cite{pang2025sga} (2025) & BS & Low to Moderate & Low to Moderate \\
GD+Sine \cite{garg2026stable} (2025) & BS & Moderate & Moderate \\
NGDiff \cite{bu2025ngdiff} (2025) & BS & Low to Moderate & Low to Moderate \\
Forgetting-MarI \cite{xu2026mari} (2025) & Approaching RA & Moderate to High & Comparable to GA \\
CATNIP \cite{yang2026catnip} (2026) & BS & Moderate & Low to Moderate \\
\bottomrule
\end{tabularx}
\end{table}

\subsection{Influence-based Gradient Methods}
\label{subsec:influence-methods}

Influence-based gradient methods aim to remove the effect of specific training data by estimating and reversing their contribution to the learned model parameters \cite{zham}, attempting to approximate how individual data points influence the final model during training \cite{jbro}.

A common formulation uses influence functions \cite{zham}, which estimate the change in model parameters when a training point is upweighted or removed. Given a model trained with parameters $\theta$, the influence of a data point $z$ on the parameters can be approximated as:
\begin{equation}
\frac{d\theta}{d\epsilon} \approx - H_\theta^{-1} \nabla_\theta \ell(z, \theta)
\label{eq:influence-function}
\end{equation}

where $H_\theta$ denotes the Hessian of the loss function with respect to model parameters and $\ell(z, \theta)$ is the loss associated with data point $z$. This expression captures how small perturbations in the training distribution affect the learned parameters. The goal is to approximate the parameter update required to remove the influence of all samples in $\mathcal{D}_f$:
\begin{equation}
\theta' \approx \theta + \frac{1}{|\mathcal{D}_f|} \sum_{z \in \mathcal{D}_f} H_\theta^{-1} \nabla_\theta \ell(z, \theta)
\label{eq:influence-update}
\end{equation}

Recent work has explored scalable approximations of this idea, using Hessian-vector products and stochastic estimation to avoid explicit computation of the Hessian inverse \cite{xfan,mckinney}. More recent efforts introduce low-rank or iterative approximations enabling partial influence removal without full retraining \cite{ak}.

However, influence functions are derived under assumptions of convexity and local smoothness that do not strictly hold for deep neural networks \cite{basu}. In highly non-convex models such as LLMs, the estimated influence of a data point may only reflect a local linearization of the loss landscape rather than its true global contribution. Furthermore, since representations are shared across many parameters and contexts, the influence of a data point cannot be cleanly isolated or inverted. Empirical findings show that even after influence-based updates, sensitive or memorized information can still be extracted under targeted prompting or adversarial evaluation \cite{patil}.

From the perspective of the proposed framework, influence-based methods aim for representation-level attenuation rather than purely behavioral suppression. By explicitly modeling the contribution of training data to parameter updates, they attempt to weaken the internal encoding of the forget set. However, due to approximation errors, non-convexity, and representational entanglement, this attenuation is often incomplete. Influence-based approaches occupy an intermediate position: they go beyond surface-level behavioral suppression by targeting parameter-level contributions, but fall short of true forgetting.

\subsection{Parameter and Loss Function Based Gradient Methods}
\label{subsec:parameter-loss-methods}

Parameter and loss function based methods approach unlearning by modifying the training objective to control how model parameters are updated during the forgetting process \cite{cheng,jbwang}. Instead of explicitly reversing gradients, these methods introduce additional penalties or constraints that limit the influence of the forget set while preserving important knowledge from the retain set.

Recent work has introduced more structured constraints on parameter updates. Methods based on remain-preserving optimization constrain updates to lie in subspaces that minimally affect retained knowledge \cite{qcheng,mluo}. LoRA-based unlearning approaches restrict updates to low-rank parameter adapters, enabling localized modification of model behavior without altering the full parameter space \cite{boer}. Parameter masking and saliency-based constraints as in SalUn \cite{fan} further refine this approach by selectively updating only parameters deemed important for the forget set.

Despite their improved control and stability, these methods rely on a critical assumption: that the knowledge associated with the forget set can be localized or approximated within a constrained subset of parameters. Empirical evidence suggests that knowledge in large language models is highly distributed and entangled across layers and contexts \cite{geva}, so restricting updates to specific subspaces may only partially remove the influence of the forget set. Constrained unlearning methods can still leak information under adversarial prompting or extraction attacks even when standard metrics indicate successful forgetting \cite{patil}.

Under the proposed framework, parameter and loss function based methods achieve a combination of behavioral suppression and partial representation-level attenuation. While they reduce the sensitivity of the model to the forget set and improve stability compared to unconstrained gradient ascent, they do not guarantee full removal of the underlying representations.

\subsection{Limitations and Failure Modes}
\label{subsec:limitations}

Despite significant progress, gradient-based unlearning methods exhibit several fundamental limitations that challenge their reliability in large language models.

A primary issue is incomplete forgetting. Due to the distributed and entangled nature of knowledge in LLMs, modifying a subset of parameters does not ensure that all traces of the target information are removed. Residual representations may persist and can be reactivated under specific prompts or contexts. Gradient-based methods are also vulnerable to adversarial recovery, as demonstrated comprehensively by Lucki et al. \cite{lucki2025adversarial}, who showed that fine-tuning an RMU-unlearned model \cite{wmdp} on as few as 10 unrelated examples restores most hazardous capabilities.

Another critical limitation is the trade-off between forgetting and utility. Strong unlearning updates, particularly those involving aggressive gradient ascent, can degrade model performance on unrelated tasks. The lack of standardized evaluation protocols makes it difficult to reliably assess unlearning effectiveness. Existing evaluation metrics such as Verbatim Probability, ROUGE-based overlap, and membership inference attacks primarily assess output-level behavior rather than internal representation removal \cite{tan}. Apparent success under these metrics does not necessarily imply true forgetting.

The over-erasure problem, identified by Xu et al. \cite{xu2026mari}, adds another dimension: existing methods remove more than they should, erasing knowledge supported by the retain set as well as by the forget set. This indiscriminate removal produces substantial utility drops and suggests that the correct unlearning objective should target only the marginal contribution of the forget data.

Collectively these limitations highlight a central challenge: current gradient-based methods may achieve apparent forgetting, but often fail to provide strong guarantees of true knowledge removal.

\section{Non-Gradient-Based Unlearning Approaches}
\label{sec:other-methods}

While gradient-based methods dominate LLM unlearning research, a rich ecosystem of non-gradient-based approaches has emerged that addresses different aspects of the forgetting problem. These methods offer complementary strengths: they avoid the optimization instability inherent in gradient ascent, can operate without access to forget-set gradients, and in some cases provide stronger formal guarantees. This section covers three principal families: model editing and knowledge editing, differential privacy based approaches, and inference-time and input modification methods.

\subsection{Model Editing and Knowledge Editing}
\label{subsec:model-editing}

Model editing, also called knowledge editing, seeks to locally modify specific factual associations within a trained LLM without retraining or performing gradient ascent over the forget set \cite{geva,memit}. The core intuition is that factual knowledge in transformer models is partially localized in feedforward MLP layers, which function as key-value memories \cite{geva}. Causal mediation analysis can identify which specific layers and neurons encode a given factual association, and targeted weight modifications can update that association without broadly affecting the rest of the model. While originally developed for knowledge updating rather than unlearning, these methods have been extensively adopted as unlearning baselines due to their ability to make targeted parametric modifications without full retraining \cite{patil}.

\textbf{ROME (Rank-One Model Editing)} \cite{30} applies this principle to perform single-fact edits using a rank-one update to the weights of a single MLP layer. The method uses causal tracing to identify the layer where the target fact is most strongly encoded, then solves a constrained optimization problem to find the minimal weight perturbation that changes the model's output for the target subject while leaving other factual associations intact. ROME is effective for single targeted edits but degrades substantially as the number of edits increases, because the independence assumptions underlying the rank-one update do not hold across multiple correlated edits.

\textbf{MEMIT (Mass-Editing Memory in a Transformer)} \cite{memit} extends ROME to support batch edits across multiple MLP layers simultaneously. Rather than targeting a single layer, MEMIT distributes the edit across a range of layers using a least-squares formulation, which improves scalability and reduces interference between simultaneous edits. MEMIT can handle thousands of edits while maintaining model utility better than ROME, and has been applied as a baseline in several LLM unlearning evaluations \cite{patil}.

\textbf{SERAC (Semi-Parametric Editing with a Retrieval-Augmented Counterfactual)} \cite{mitchell2022serac} takes a fundamentally different approach: rather than modifying model weights directly, SERAC maintains a small external memory of edits and routes queries through it at inference time. When a query matches an edited fact, the counterfactual model overrides the base model's response. This avoids weight modification entirely, preserving model utility, but requires storing the forget data and does not achieve parametric removal.

\textbf{MEND (Model Editor Networks with Gradient Decomposition)} \cite{mitchell2022mend} applies low-rank transformations to efficiently apply edits via auxiliary networks trained to predict the optimal weight update for a given edit. MEND is more computationally efficient than ROME and MEMIT for repeated edits but similarly operates at the behavioral level.

From the perspective of the three-level framework, model editing methods achieve behavioral suppression for the specific facts they target. They do not achieve representation-level attenuation because causal tracing identifies the most salient encoding location of a fact, not all locations where the fact's influence is distributed across parameters. Patil et al. \cite{patil} showed empirically that content deleted via ROME and MEMIT remains accessible in hidden states and via rephrased queries, highlighting that model editing faces the same parametric retention challenge as gradient-based methods.

A growing body of work directly compares model editing to unlearning baselines. \textbf{WISE (Working memory In Side-memory Editing)} \cite{wang2024wise} addresses the limitations of single-layer editing by introducing a dual parametric memory scheme, maintaining a main memory for pretrained knowledge and a side memory for edited knowledge, with a learned router directing queries to the appropriate memory. This approach overcomes the impossible triangle of reliability, generalization, and locality that afflicts prior editing methods under sequential editing. \textbf{AlphaEdit} \cite{fang2024alphaedit} projects parameter perturbations onto the null space of the preserved knowledge before applying them, theoretically guaranteeing that the output of the post-edited model remains unchanged when queried about non-targeted knowledge, and achieves an average 36.7\% performance improvement over standard locating-then-editing methods. \textbf{PMET (Precise Model Editing in a Transformer)} \cite{li2023pmet} improves upon ROME and MEMIT by simultaneously optimizing hidden states of both the Multi-Head Self-Attention and Feed-Forward Network components, but updating only the FFN weights, producing more precise factual edits. Both WISE and AlphaEdit have been found effective as unlearning baselines, particularly for pretrained knowledge, and excel in generating human-aligned refusal responses. The relationship between knowledge editing and unlearning motivates a hybrid perspective: editing identifies where knowledge is localized, while unlearning determines how to remove it. Future methods combining causal tracing for localization with gradient-based updates for removal may offer stronger forgetting guarantees than either approach alone.

\subsection{Differential Privacy Based Approaches}
\label{subsec:differential-privacy}

Differential privacy (DP) provides a mathematical framework for quantifying and bounding the influence of any individual data point on a trained model. A mechanism $M$ satisfies $(\epsilon, \delta)$-differential privacy if for any two datasets $\mathcal{D}$ and $\mathcal{D}'$ differing in exactly one data point, and for any measurable output set $\mathcal{S}$:
\begin{equation}
\Pr(M(\mathcal{D}) \in \mathcal{S}) \leq e^\epsilon \Pr(M(\mathcal{D}') \in \mathcal{S}) + \delta
\label{eq:differential-privacy}
\end{equation}

The connection between differential privacy and machine unlearning is direct: if a model is trained with $(\epsilon, \delta)$-DP guarantees, then the deletion capacity of that model can be bounded formally \cite{cao2015towards}. This provides the theoretical grounding that approximate unlearning methods based on DP offer, which gradient-only methods lack.

\textbf{DP-SGD (Differentially Private Stochastic Gradient Descent)} \cite{abadi2016dpsgd} is the canonical DP training procedure, included here as the foundational mechanism underlying recent DP-based unlearning work. It clips per-example gradients to a maximum norm $C$ and adds calibrated Gaussian noise to the aggregated gradient before each parameter update. The clipping ensures that no single training example can shift the gradient by more than $C$, and the noise masks the contribution of any individual example. Models trained with DP-SGD have bounded sensitivity to any single training example, which directly supports approximate unlearning: removing a data point from a DP-trained model requires only a bounded correction rather than full retraining.

\textbf{DP2Unlearning} \cite{dp2unlearning2025} combines DP with post-hoc unlearning mechanisms to provide formal forgetting guarantees for LLMs that existing gradient-based methods cannot offer. The approach addresses a fundamental limitation of approximate unlearning methods: they rely on empirical evidence rather than formal guarantees, which fails to satisfy the Right to Be Forgotten as stated in GDPR and similar legal frameworks. By leveraging DP's formal privacy bounds, DP2Unlearning can certify that the influence of a removed data point falls below a quantifiable threshold.

The primary limitation of DP-based approaches is the privacy-utility trade-off: the noise added to achieve DP guarantees degrades model utility, particularly for LLMs trained on large and diverse corpora. Furthermore, DP training is a preventive control: it must be applied during training and is not retroactive. A deployed LLM that was not trained with DP cannot retroactively acquire DP guarantees, which limits the applicability of DP-based approaches to models in development rather than already-deployed systems.

Despite these limitations, DP-based methods represent the only family of approaches that provide formal, mathematically verifiable forgetting guarantees. As regulatory pressure for demonstrable compliance with erasure rights increases, DP-based unlearning may become increasingly important for deployment contexts where formal certification is required.

\subsection{Inference-Time and Input Modification Methods}
\label{subsec:inference-methods}

A distinct class of unlearning approaches operates at inference time rather than modifying model weights. These methods simulate unlearning by altering the model's input or output pipeline, making them applicable to black-box models where weight access is unavailable.

\textbf{In-context unlearning (ICUL)} \cite{pawelczyk2023icul} modifies the model's input context to signal that certain knowledge should not be expressed. Label flipping disrupts the original association by providing counter-examples in the context window, effectively overriding the model's parametric knowledge with in-context instructions. ICUL requires no weight modification and is applicable to black-box APIs, but provides no guarantee of parametric removal and requires storing the forget-set content for prompt construction, which may contradict the data privacy goals motivating unlearning.

\textbf{Soft prompt optimization} learns continuous input representations that, when prepended to queries about the forget set, induce the model to produce alternative or refusal responses. Methods such as SPUL \cite{bhaila2024spul} optimize soft prompt tokens to selectively suppress knowledge without modifying model weights. These approaches are efficient and reversible, but their effectiveness is limited to the distribution of queries for which the soft prompt was optimized.

\textbf{Logit-offset methods} operate on model outputs at decoding time. Methods such as $\delta$-Unlearning \cite{ji2024uld} compute a logit offset between a base model and a forgetting model, subtracting the forgetting signal from the base model's output distribution at inference. ULD (Unlearning by Logit Difference) \cite{ji2024uld} subtracts logits from an auxiliary model trained with reversed objectives. These methods require access to model logits and offer no parametric removal guarantee, but can be applied to already-deployed models without retraining.

Inference-time methods occupy the behavioral suppression tier of the three-level framework. Their principal advantage is applicability to black-box or nearly-black-box settings where weight modification is impossible. For regulatory compliance purposes, their inability to demonstrate parametric removal is a significant limitation.

Table~\ref{tab:nongradient-methods} summarizes all non-gradient-based methods covered in this section across the dimensions of the framework.

\begin{table}[h]
\caption{Comparison of non-gradient-based unlearning methods. Forgetting level follows the framework in Section~\ref{subsec:framework}. Weight access indicates the level of model access required at unlearning time. Formal guarantee indicates whether the method provides a mathematically certifiable forgetting bound. Foundational methods predating 2023 are included where they directly underpin recent unlearning research. BS = Behavioral Suppression, RA = Representation Attenuation (approaching).}
\label{tab:nongradient-methods}
\centering
\renewcommand{\arraystretch}{1.25}
\setlength{\tabcolsep}{4pt}
\begin{tabularx}{\linewidth}{>{\raggedright\arraybackslash}X
                              >{\centering\arraybackslash}p{0.9cm}
                              >{\raggedright\arraybackslash}p{2.0cm}
                              >{\raggedright\arraybackslash}X
                              >{\centering\arraybackslash}p{1.35cm}}
\toprule
\textbf{Method} & \textbf{Year} & \textbf{Forgetting Level} & \textbf{Weight Access} & \textbf{\shortstack{Formal\\Guarantee}} \\
\midrule
DP-SGD \cite{abadi2016dpsgd}           & 2016 & Approaching RA & Yes (training-time) & Yes \\
ROME \cite{30}                         & 2022 & BS             & Yes (full)          & No  \\
MEMIT \cite{memit}                     & 2022 & BS             & Yes (full)          & No  \\
SERAC \cite{mitchell2022serac}         & 2022 & BS             & No                  & No  \\
MEND \cite{mitchell2022mend}           & 2022 & BS             & Yes (full)          & No  \\
ICUL \cite{pawelczyk2023icul}          & 2024 & BS             & No                  & No  \\
ULD \cite{ji2024uld}                   & 2024 & BS             & Partial (logits)    & No  \\
DP2Unlearning \cite{dp2unlearning2025} & 2025 & Approaching RA & Yes (training-time) & Yes \\
SPUL \cite{bhaila2024spul}             & 2025 & BS             & No                  & No  \\
\bottomrule
\end{tabularx}
\end{table}

\subsection{Limitations of Non-Gradient Approaches}
\label{subsec:nongradient-limitations}

Non-gradient-based unlearning approaches share several fundamental limitations. Model editing methods achieve behavioral modification for targeted facts but leave the knowledge distributed across other parameter regions intact, making them susceptible to the same extraction attacks that afflict gradient-based methods. DP-based approaches provide formal guarantees but require DP training from the outset and impose significant utility costs that are challenging to manage at LLM scale. Inference-time methods provide no parametric removal and are effective only within the distribution of queries for which they were designed.

Across all three families, a consistent pattern emerges that mirrors the gradient-based case: the depth of removal achieved is behavioral suppression rather than parametric elimination. No non-gradient-based method reviewed here reliably achieves representation-level attenuation, let alone true forgetting. The field-wide challenge of moving from suppression to genuine forgetting applies equally to all current method families.

\section{Evaluation of Unlearning}
\label{sec:evaluation}

Evaluating unlearning in large language models remains a fundamentally challenging problem due to model scale, generative behavior, and the absence of a tractable ground truth \cite{wu}. Unlike classical machine unlearning, where the gold standard is comparison with a model retrained from scratch without the forget set \cite{tang,lag}, such retraining is computationally infeasible for modern LLMs. Existing evaluation strategies therefore rely on proxy metrics and benchmark datasets that assess different aspects of unlearning rather than verifying true removal of internal representations.

\subsection{Evaluation Protocols and Baselines}
\label{subsec:evaluation-protocols}

Most LLM unlearning studies adopt a standardized evaluation pipeline in which a pretrained model is fine-tuned on controlled datasets followed by targeted unlearning \cite{lizzo}. The resulting model is compared against baseline methods such as gradient ascent, gradient difference, or KL-regularized variants. The lack of a unified evaluation standard leads to inconsistencies across studies, making direct comparison difficult.

\subsection{Benchmark Datasets}
\label{subsec:benchmark-datasets}

\subsubsection{Unlearning-Specific Benchmarks}

Several benchmarks have been proposed to explicitly evaluate unlearning. TOFU (Task of Fictitious Unlearning) \cite{tofu} introduces synthetic entities, enabling controlled evaluation where the ground truth is known, with separate forget, retain, and generalization sets. RWKU (Real-World Knowledge Unlearning) \cite{rwku} focuses on real-world knowledge and incorporates additional subsets for membership inference attacks and utility evaluation. WMDP (Weapons of Mass Destruction Proxy) \cite{wmdp} targets hazardous knowledge domains such as biosecurity and cybersecurity, evaluating whether models can be prevented from generating harmful information. The Harry Potter Benchmark \cite{harry} evaluates memorization and removal of domain-specific knowledge through prompt-based testing. These benchmarks enable controlled evaluation of forgetting effectiveness, though they primarily assess output-level behavior.

Table~\ref{tab:benchmarks} provides a concise comparison of the major unlearning benchmarks.

\begin{table}[h]
\caption{Comparison of major LLM unlearning benchmarks. Ground truth indicates whether the forget set is fully separable from the retain set. Adversarial evaluation indicates whether the benchmark includes robustness testing against recovery attacks.}
\label{tab:benchmarks}
\centering
\renewcommand{\arraystretch}{1.25}
\setlength{\tabcolsep}{4pt}
\begin{tabularx}{\linewidth}{>{\raggedright\arraybackslash}X
                              >{\centering\arraybackslash}p{0.9cm}
                              >{\raggedright\arraybackslash}X
                              >{\centering\arraybackslash}p{1.6cm}
                              >{\centering\arraybackslash}p{1.6cm}}
\toprule
\textbf{Benchmark} & \textbf{Year} & \textbf{Forget Set Type} & \textbf{Ground Truth} & \textbf{Adversarial Evaluation} \\
\midrule
Harry Potter \cite{harry} & 2023 & Copyrighted domain text & No & No \\
TOFU \cite{tofu}          & 2024 & Synthetic (fictitious authors) & Yes & No \\
RWKU \cite{rwku}          & 2024 & Real-world knowledge & Partial & Yes (MIA) \\
WMDP \cite{wmdp}          & 2024 & Hazardous knowledge & No & No \\
MUSE \cite{muse2024}      & 2024 & Multi-domain (copyright, news) & No & Yes \\
\bottomrule
\end{tabularx}
\end{table}

\subsubsection{General Capability Benchmarks}

To evaluate utility preservation, prior work commonly relies on MMLU \cite{mmlu}, TruthfulQA \cite{tqa}, HellaSwag \cite{swag}, and ARC \cite{arc}. These benchmarks collectively assess a model's performance across factual knowledge recall, commonsense reasoning, and language understanding. A significant drop in performance after unlearning indicates interference with retained knowledge, while stable performance suggests effective preservation of utility.

\subsection{Evaluation Metrics}
\label{subsec:evaluation-metrics}

Forgetting effectiveness is typically evaluated using ROUGE-L overlap between generated and target content, the log-likelihood or probability of generating forget-set data, and measures like truth ratio and perplexity. These metrics assess whether the model avoids reproducing information associated with the forget set, but primarily capture output-level behavior and do not guarantee that the underlying knowledge has been fully removed.

Utility preservation is assessed by measuring model performance on the retain set and on standard benchmarks. Efficiency is evaluated in terms of computational cost relative to retraining. The NeurIPS 2023 Unlearning Challenge, for example, considers methods efficient if their computational cost is below 20\% of the retraining time \cite{tri}.

A recently identified limitation of classical metrics is the squeezing effect \cite{li2026beliefs}: methods that score well on ROUGE and perplexity may still leak targeted knowledge in paraphrased form. This motivated the development of LLM-as-a-judge evaluation as a semantic complement to surface-level metrics.

\subsection{Robustness and Adversarial Evaluation}
\label{subsec:robustness-adversarial}

A critical aspect of unlearning evaluation is robustness against recovery attacks. Membership Inference Attacks (MIA) are commonly used to determine whether traces of the forget set remain in the model \cite{liu}, where a high attack success rate indicates incomplete unlearning. Relearning attacks demonstrate that supposedly forgotten knowledge can be rapidly recovered through fine-tuning on small auxiliary datasets \cite{blanco}. Adversarial prompting techniques including Dynamic Unlearning Attacks \cite{yuan} and jailbreak-style prompts attempt to elicit forgotten information through carefully optimized inputs. Embedding-space attacks using soft prompting operate directly on latent representations to recover hidden knowledge \cite{schwinn}.

The most comprehensive adversarial evaluation of state-of-the-art unlearning methods is provided by Lucki et al. \cite{lucki2025adversarial}, who systematically apply white-box and gray-box attacks to models unlearned on the WMDP benchmark. Fine-tuning an RMU-unlearned model \cite{wmdp} on as few as 10 unrelated examples restores most hazardous capabilities. Orthogonalization of residual stream directions recovers WMDP-Biology accuracy to 64.7\% from the post-unlearning level, equaling the pre-unlearning baseline of 64.4\%, without modifying any model weights. The paper concludes that existing unlearning methods are not meaningfully different from safety fine-tuning in terms of robustness: both primarily obfuscate knowledge rather than removing it.

The robustness problem extends beyond knowledge recovery to the retain set itself. Dang et al. \cite{dang2026rna} identify retain-robustness as a failure mode distinct from forget-set knowledge recovery: unlearning methods inadvertently train the model to treat forget-set tokens as backdoor triggers, so that when those tokens appear in otherwise benign retain-set queries, the model's responses are degraded. The proposed mitigation, Random Noise Augmentation applied during the retain phase, significantly improves retain-robustness while preserving forget-set performance.

Fan et al. \cite{fan2025sam} establish a connection between LLM unlearning robustness and sharpness-aware minimization. Standard NPO places the unlearned model at a sharp local minimum in the loss landscape, making small weight perturbations of the kind applied in a relearning attack sufficient to reverse the unlearning. The robust unlearning objective is formulated as the min-max problem:
\begin{equation}
\min_{\theta}\, \max_{\|\delta\|_p \leq \rho}\; \ell_f(\theta + \delta \mid \mathcal{D}_f) + \lambda\,\ell_r(\theta \mid \mathcal{D}_r)
\label{eq:sam-minmax}
\end{equation}

where the inner maximization simulates the worst-case weight perturbation a relearning attacker would apply. This aligns with sharpness-aware minimization, which promotes a flat loss landscape around the unlearned solution. Solving the inner maximization in closed form via linear approximation yields the SAM forget loss:
\begin{equation}
\ell_f^{\text{SAM}}(\theta) = \ell_f\!\left(\theta + \rho\,\frac{\nabla_\theta \ell_f(\theta)}{\|\nabla_\theta \ell_f(\theta)\|_2}\right)
\label{eq:sam-loss}
\end{equation}

NPO+SAM is reported to match the robustness of tamper-resistant safeguards (TAR) \cite{tamirisa2024tar}, which uses computationally intensive meta-learning, while running over 600 times faster on equivalent tasks \cite{fan2025sam}. Smoothness-based robustness also extends to jailbreaking attacks, not only relearning attacks.

Rezkellah and Dakhmouche \cite{rezkellah2025constrained} propose a unified framework for both unlearning and jailbreak robustness through constrained optimization directly on model weights. The Point-wise Constrained Regions (PCR) method identifies forbidden concept embeddings and applies the minimal weight perturbation that forces MLP layer outputs to remain geometrically distant from those embeddings, using Karush-Kuhn-Tucker (KKT) conditions for the constrained optimization in each layer. Across Llama-3.1-8B, Gemma-2B-IT, and Mistral-7B, PCR achieves refusal rates of 100\%, 97.4\%, and 26.7\% respectively against projected gradient descent jailbreak attacks, compared to 87.5\%, 10\%, and 37.5\% for SmoothLLM \cite{robey2023smoothllm} on the same models.

\subsection{Discussion: Limitations of Current Evaluation}
\label{subsec:eval-limitations}

Current evaluation methods suffer from a fundamental limitation: they predominantly assess behavioral suppression rather than true forgetting. Most metrics rely on observing model outputs under predefined prompts, which only capture whether the model avoids generating specific responses. As demonstrated by adversarial and relearning attacks, suppressed knowledge can often be recovered, indicating that it remains encoded in model parameters. Existing evaluation protocols are well-suited for detecting behavioral suppression, but insufficient for verifying representation-level attenuation or true forgetting. Developing evaluation methods that directly probe internal representations or provide formal guarantees remains an open challenge.

\section{Discussion and Open Challenges}
\label{sec:discussion}

The central finding of this survey, that current gradient-based unlearning methods achieve behavioral suppression rather than true forgetting, has direct implications for their deployment as security controls. A behavioral suppression defense is one whose guarantees hold only for the class of adversary assumed during evaluation. In the majority of published evaluations, that adversary is black-box, non-adaptive, and operating under standard prompting conditions. The adversarial evaluations reviewed in Section~\ref{subsec:robustness-adversarial} demonstrate that these guarantees dissolve under gray-box and white-box access. The failure of current unlearning to function as a robust security control is not a weakness of any specific method but a structural property of the optimization approach: gradient-based updates that modify output behavior without altering internal representations leave intact the parameter-level pathways through which a capable adversary can recover the target knowledge.

Current LLM unlearning methods largely conflate two distinct objectives: suppressing the observable expression of knowledge and removing the underlying representations that encode it. Most existing gradient-based approaches primarily achieve the former, while offering limited evidence of the latter. In deployed systems, successful suppression under standard prompts is often interpreted as successful data removal. Prior work shows that suppressed information can frequently be recovered through adversarial prompting, relearning attacks, or latent-space probing \cite{schwinn,blanco,yuan}. This gap raises concerns about whether current unlearning practices provide meaningful guarantees in privacy-sensitive or safety-critical applications.

Current evaluation protocols primarily measure output behavior under fixed prompting distributions, making them well-aligned with detecting behavioral suppression but insufficient for assessing representational removal. Adversarial evaluation methods including membership inference, relearning attacks, and prompt-based extraction partially address this gap by probing recovery pathways, but they test specific access strategies rather than providing guarantees over all possible reconstruction mechanisms. This creates a structural evaluation gap: current benchmarks can confirm when unlearning fails, but cannot certify when it succeeds in the strong sense of true forgetting.

\subsection{Limitations of Gradient-Based Methods}
\label{subsec:gradient-limitations}

The limitations of gradient-based unlearning arise from a structural mismatch between the optimization mechanism and how knowledge is encoded in large language models. Training in LLMs is a highly non-convex, path-dependent process in which information becomes distributed and entangled across parameters \cite{geva}. Gradient ascent on a forget set therefore does not invert training dynamics; it induces a new trajectory that modifies output behavior without reliably isolating or eliminating the associated internal representations.

This limitation is further amplified by parameter sharing: the same weights that encode forget-set information also support unrelated capabilities. Stronger updates risk degrading general utility, producing the well-known trade-off between forgetting effectiveness and model performance \cite{guo2019certified,anjarlekar2025grin}. Influence-based methods attempt to approximate training contributions using second-order information, but rely on local linearity assumptions that do not hold in large-scale non-convex settings \cite{basu}. Parameter-efficient and saliency-based approaches improve update precision but still assume partial localizability of knowledge, reducing sensitivity to targeted data without guaranteeing removal of the underlying representations.

\subsection{What the Framework Reveals}
\label{subsec:framework-reveals}

Mapping current approaches onto the behavioral suppression, representation attenuation, true forgetting hierarchy reveals a consistent pattern: most gradient-based and constrained optimization methods operate at the level of behavioral suppression, while influence-based techniques occasionally achieve partial representational attenuation. No existing method reliably achieves true forgetting in the sense of equivalence to retraining without the forget set. This mapping clarifies an important conceptual point: similar performance on existing benchmarks does not imply equivalent levels of forgetting. A method that suppresses outputs under standard prompts may still retain recoverable representations internally, meaning that evaluation results can overestimate the degree of actual information removal. Improvements in benchmark performance should therefore be understood as improvements in suppression capability rather than guarantees of knowledge deletion.

\subsection{Open Challenges}
\label{subsec:open-challenges}

Despite progress in LLM unlearning, several fundamental challenges remain unresolved.

\textbf{Challenge 1: Achieving and verifying true forgetting.} Current methods largely operate by suppressing behavioral outputs rather than removing internal representations, and there is still no reliable way to verify whether knowledge has been genuinely eliminated. Output-based evaluation is insufficient, while direct inspection of representations remains an open problem.

\textbf{Challenge 2: Robustness to adversarial recovery.} Existing unlearning methods can often be bypassed through adversarial prompting, relearning attacks, or latent-space extraction techniques \cite{yuan,blanco,schwinn}. No current method provides robustness guarantees against adaptive white-box adversaries.

\textbf{Challenge 3: Scalability and compositionality.} Although parameter-efficient and gradient-based methods reduce cost compared to retraining, they remain expensive at scale and are not well-suited for continual or repeated unlearning. Real-world knowledge is highly interconnected, making fine-grained or compositional removal of specific facts difficult without unintended side effects.

\textbf{Challenge 4: Adversarially adaptive evaluation.} Current unlearning benchmarks including TOFU \cite{tofu}, RWKU \cite{rwku}, WMDP \cite{wmdp}, and MUSE \cite{muse2024} test model behavior under predefined and non-adaptive prompting conditions. An adversary with knowledge of the unlearning procedure applied can craft adaptive attacks that systematically search for conditions under which unlearned knowledge resurfaces. No standardized benchmark currently tests unlearned models against adaptive white-box adversaries, meaning that existing evaluation results cannot certify robustness against a determined adversary with model access. Developing such evaluation frameworks is a prerequisite for deploying unlearning as a credible security control in high-stakes settings.

\textbf{Challenge 5: Theoretical and evaluative foundations.} Current methods are largely heuristic, with limited understanding of when or why unlearning approximations work. There is no standardized evaluation framework that reliably distinguishes behavioral suppression from genuine forgetting, making comparisons across methods inconsistent. Bridging this gap is essential for both scientific progress and alignment with regulatory requirements such as the right to erasure under GDPR Article 17 and similar data deletion laws.

\section{Conclusion}
\label{sec:conclusion}

This survey has examined LLM unlearning through the lens of cybersecurity, robustness, and verifiable forgetting. The central conclusions are as follows.

\begin{itemize}
\item \textbf{Behavioral suppression is not security.} The vast majority of current gradient-based unlearning methods achieve behavioral suppression under standard prompting conditions. They do not achieve representation-level attenuation or true forgetting, and their guarantees dissolve under gray-box or white-box adversarial access.

\item \textbf{The evaluation gap is a security audit gap.} Current benchmarks assess behavioral suppression but cannot certify parametric removal. This is not merely a methodological limitation; it is a fundamental barrier to deploying unlearning as a verifiable security control.

\item \textbf{The three-level framework provides a diagnostic tool.} Classifying methods under behavioral suppression, representation-level attenuation, and true forgetting gives practitioners a clear vocabulary for specifying what level of security guarantee a deployment context requires and whether any available method can provide it.

\item \textbf{Non-gradient approaches complement but do not solve the core problem.} Model editing methods achieve targeted behavioral modification, DP-based approaches provide formal guarantees for models trained with DP from the outset, and inference-time methods extend unlearning to black-box settings. None of these approaches reliably achieves true parametric forgetting.

\item \textbf{The field requires adversarially adaptive evaluation.} No standardized benchmark currently tests unlearned models against adaptive white-box adversaries. Developing such evaluation frameworks is the most critical infrastructure need in the field.
\end{itemize}

LLM unlearning is a rapidly growing field with direct implications for regulatory compliance, safety-critical deployment, and the long-term trustworthiness of AI systems. The gap between the field's current capabilities and the security guarantees that high-stakes deployments require is large, but it is precisely defined. Closing it requires methods that operate at the level of internal representations rather than output distributions, evaluation protocols that assume capable adversaries rather than passive users, and formal frameworks that can certify the absence of parametric knowledge rather than merely its behavioral suppression.

\section{Author Contributions}
 R. S. Shankar performed the literature review, data collection and curation, methodology and software design, investigation, and manuscript drafting. A. Bhardwaj performed the literature review, data collection and curation, methodology and software design, investigation, and manuscript drafting. A. Doshi performed content validation, manuscript review, manuscript editing, and manuscript presentation.
 A. Nagarajan performed content validation, manuscript review, and manuscript editing.
T. P. Asia performed content validation, manuscript review, and manuscript editing.
 S. Sengupta performed project administration, supervision, methodological guidance, content validation, manuscript review, and manuscript editing.

\section{Funding}
This research did not receive external funding.

\section{Data Availability Statement}
All datasets analyzed in this survey are publicly available. No new data were generated for this study.

\section{Declaration of Competing Interest}
The authors declare no competing interests.

\bibliographystyle{unsrtnat}
\bibliography{cas-refs}

\end{document}